\documentclass[a4paper]{article}

\usepackage{INTERSPEECH2021}

\usepackage{times}
\usepackage{latexsym}
\usepackage{booktabs}
\usepackage{float}
\usepackage[disable]{todonotes}
\usepackage{svg}

\usepackage{hyperref} 

\title{On-the-Fly Aligned Data Augmentation for Sequence-to-Sequence ASR}

\name{Tsz Kin Lam$^1$, Mayumi Ohta$^1$, Shigehiko Schamoni$^{1,2}$, Stefan Riezler$^{1,2}$}

\address{
  $^1$Computational Linguistics \& 
  $^2$IWR, Heidelberg University, Germany}
\email{\{lam,ohta,schamoni,riezler\}@cl.uni-heidelberg.de}

\begin{document}

\maketitle
\begin{abstract}
We propose an on-the-fly data augmentation method for automatic speech recognition (ASR) that uses alignment information to generate effective training samples. 
Our method, called Aligned Data Augmentation (ADA) for ASR, replaces transcribed tokens and the speech representations in an aligned manner to generate previously unseen training pairs. The speech representations are sampled from an audio dictionary that has been extracted from the training corpus and inject speaker variations into the training examples. The transcribed tokens are either predicted by a language model such that the 
augmented data pairs are semantically close to the original data, or randomly sampled. Both strategies result in training pairs that improve robustness in ASR training. 
Our experiments on a Seq-to-Seq architecture show that ADA can be applied on top of SpecAugment, and achieves about 9--23\% and 4--15\% relative improvements in WER over SpecAugment alone on LibriSpeech 100h and LibriSpeech 960h test datasets, respectively.
\end{abstract}

\section{Introduction}
Data augmentation is a common technique used in training machine learning models. It improves model performance by increasing the amount of training data and by reducing overfitting at the same time, thus increasing overall model robustness. In computer vision, common approaches apply geometric transformations or add noise to images to make the models more robust \cite{lecun98gradient-basedlearning,simard2003best}. 
In machine translation, a straightforward idea is to replace tokens on both source and target texts \cite{wang2018switchout,fadaee-etal-2017-data}, or, more complex, apply back-translation using a reverse translation model to generate additional training examples \cite{sennrich2016improving}. 

In automatic speech recognition (ASR), data augmentation methods range from simple perturbations on the audio inputs, e.g. changing speed \cite{ko2015audio} or dynamic time stretching \cite{nguyen2020improving}, or masking out sections \cite{Park2019}, to more complex techniques such as pseudo labelling \cite{kahn2020selftraining}, sampling noisy counterparts to examples \cite{liang-etal-2018-noisysampling}, 
utilizing multi-modal information \cite{renduchintala-2018},
or combining a Text-To-Speech (TTS) engine with a speech synthesizer \cite{laptev2020you,rossenbach-etal-2020-synthetic,du-etal-2021-dataaug} to generate training data from transcriptions. 
Computationally simpler methods usually augment data on-the-fly, while the more complex methods require considerable resources, 
which makes them unsuitable for on-the-fly data augmentation.

In this work, we propose a novel data augmentation method which is able to apply variations on both source audio and target text to generate new training examples in a computationally efficient manner. 
The central component of our approach is an audio dictionary that allows us to apply our strategies to data augmentation on-the-fly, thus increasing variability in contrast to offline augmentation, and improving efficiency by avoiding the necessity to store and load augmented data.

On the target text, we either use a masked language model to replace tokens with semantically close variations, or we replace tokens following a random token strategy. In the former case, the augmented text differs not too much from the original text, while in the latter case, the possibly ungrammatical examples force the model to put more focus on the audio input and less focus on the inherent language model. Both approaches successfully increase robustness, i.e. performance on previously unseen examples \cite{ng2020ssmba}.

On the source audio, sections are replaced by entries sampled from an audio dictionary. Our audio dictionary is extracted from the training corpus and its functionality is comparable to the aforementioned TTS+speech synthesizer method, Vocal Tract Length Perturbation \cite{jaitly-hinton-2015-vtlp}, or Stochastic Feature Mapping \cite{cui-etal-2015-augmentation}, however, it requires much less computing power and the audio representation replacements are from real human speech. 
The resulting audio sequence is thus a combination of the original audio representations and replacements, which is similar to data augmentation techniques for images such as CutMix \cite{yun2019cutmix}. 

The novelty of our method, called Aligned Data Augmentation (ADA) for ASR, is to use alignment information to produce new parallel data by replacing parts of target text and source audio in a synchronized manner to generate previously unseen data. The additional training examples created by our method improve over one-sided replacements, even in combination with existing data augmentation techniques like SpecAugment \cite{Park2019}.

Recently, other works showed the utility of alignment information for training of speech-to-text systems. 
\cite{salesky2019exploring} improve end-to-end speech translation by using phoneme-level alignments. 
However, they use this information to compress phoneme representations and they do not increase the amount of training data. 
\cite{nguyen2020improving} propose to create subsequences by truncating source audio and target transcriptions in an aligned manner. 
In contrast to this, we generate complete previously unseen examples by partial replacements of source audio and target text.

One crucial difference between ADA and other semi-supervised learning methods such as \cite{kahn2020selftraining,laptev2020you,rossenbach-etal-2020-synthetic} is that ADA uses the original dataset, while the other three include additional in-domain unlabelled audio or text to generate augmented data pairs. Furthermore, the on-the-fly data augmentation done in ADA differentiates our technique from complex techniques that require offline preprocessing, e.g. speed perturbations \cite{ko2015audio}. 

A recent, very sophisticated online augmentation method, Sample-Adaptive Policy for Augmentation \cite{hu2020sapaugment}, showed huge gains on the LibriSpeech datasets. They learn a policy that combines augmentation methods based on the training loss of data examples. We combine different augmentation methods and we employ a simple static mixture schedule. Despite the simplicity of our method, we achieve improvements comparable to theirs.

\begin{figure}[ht!]
    \centering
    \includegraphics[width=0.47\textwidth]{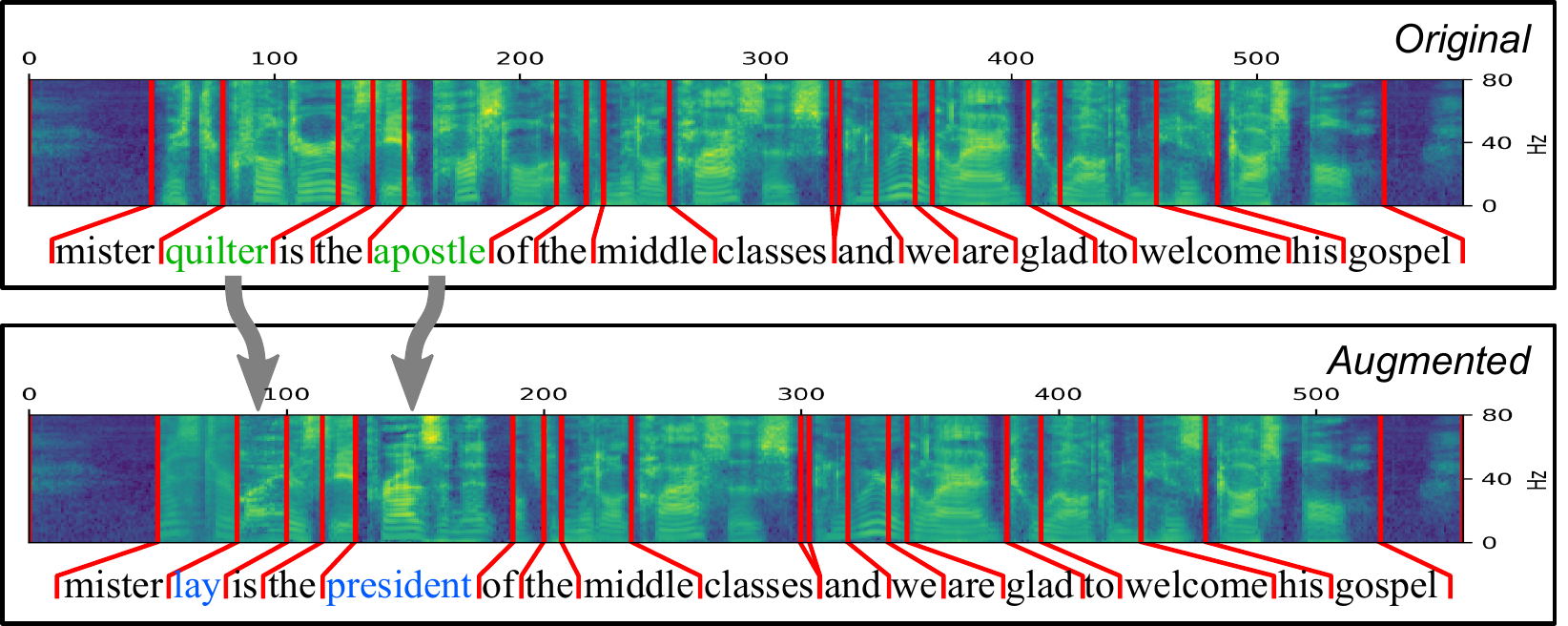}\\
    \caption{Example from the LibriSpeech dataset illustrating aligned data augmentation: In the original audio-text pair, certain tokens (green) are replaced following certain strategies (blue). An audio dictionary created on the training data is then queried to replace the aligned audio representations of the predicted tokens, resulting in an augmented audio-text pair. 
    }
    \label{fig:ada_example}
    \vspace{-4mm}
\end{figure}

\section{Aligned Data Augmentation}
Figure \ref{fig:ada_example} illustrates the ADA process of creating an augmented data pair from an original data pair. Starting with the target side of the original pair, we randomly select tokens for replacement (printed in green). 
These tokens are then replaced on the augmented target side by candidate tokens (printed in blue) following two alternative strategies: (1) language model guided strategy which we call ADA-LM, and (2) random token strategy which we call ADA-RT. 
Finally, an audio dictionary is queried for audio representations corresponding to the replaced tokens, resulting in a new augmented data pair. 
In case the replacement token suggested by the language model has no corresponding entry in the audio dictionary which happens in less than 5\% of the replacements, ADA-LM masks out the aligned audio representation. Data pairs augmented in that manner look very similar to the ones generated by SpecAugment, however, we inject additional aligned target side variations. 
The candidates suggested by the language model are in general semantically close, such as the replacement of ``apostle'' with ``president''.
By exchanging the names ``quilter'' and ``lay'', as suggested by the language model, ADA-LM receives out-of-domain knowledge, increasing model robustness. 
The second strategy, ADA-RT, samples a random token and then replaces the corresponding audio representation. This mostly results in ungrammatical sentences, however, it forces the model to adjust the influence of the inherent language model. 

For each token, our audio dictionary usually keeps multiple audio representations available that differ in tone, speed, or speaker. 
Such an audio dictionary introduces several aspects useful for ASR training. First, it leads to clean augmented data pairs containing real human speech. Then, it introduces variations of prosody and gender in the audio representations. Finally, it is computationally simple and can be applied on-the-fly.

\subsection{Language Model Only} 
In a separate experiment, we evaluate the impact of utilizing only a language model to generate new data pairs. In this scenario, we do not mask out the corresponding audio segments and only apply SpecAugment to the audio representations. This effectively creates variations in the transcriptions that do not perfectly match with the audio representations. 

\subsection{Audio Dictionary Only} The audio dictionary can also be applied without text token replacement. Here, the source audio representation of a word that is not changed on the target side is replaced by a sample of the same token from the dictionary. Since ASR is a many-to-one problem, audio-text pairs modified in this manner increase the recognition performance. This happens for the word ``mister'' in Figure \ref{fig:ada_example} where the audio representation in the augmented pair is replaced by an entry from the dictionary. In our experiments in Section \ref{sec:experiments}, we use this source-side-only replacement to evaluate the contribution of the audio dictionary. A source-side-only replacement can also occur rarely in the aligned case if the language model predicts the same token as the original one, e.g. when function words or parts of common named entities are to be replaced by the language model.

\begin{table*}[ht]
\centering
\footnotesize
{
\begin{tabular}{llcrrrrr}
\toprule
\# & model & type & \texttt{dev-clean} & \texttt{dev-other} & \texttt{test-clean} & \texttt{test-other} & \texttt{time} \\
\midrule
1 & SpecAugment (baseline) & -- & $10.51$ $\pm$ $0.04$ & $22.92$ $\pm$ $0.17$ & $11.50$ $\pm$ $0.10$ & $23.50$ $\pm$ $0.11$  & $1.0$ \\
2 & LanguageModel & target only & ${^{1}}9.98$ $\pm$ $0.13$ & ${^{1}}22.31$ $\pm$ $0.19$  & ${^{1}}10.88$ $\pm$ $0.15$ & ${^{1}}22.80$ $\pm$ $0.15$ & $2.0$ \\
3 & AudioDict & source only & ${^{1}}9.90$ $\pm$ $0.05$ & ${^{1,2}}21.50$ $\pm$ $0.10$  & ${^{1}}10.70$ $\pm$ $0.04 $ & ${^{1,2}}22.12$ $\pm$ $0.09$ & $1.2$ \\
4 & ADA-LM & aligned & ${^{1,2,3}}9.26$ $\pm$ $0.07$ & ${^{1,2}}21.30$ $\pm$ $0.11$  & ${^{1,2,3}}10.12$ $\pm$ $0.06$  & ${^{1,2,3}}21.41$ $\pm$ $0.08$ & $2.4$ \\
5 & ADA-RT & aligned & ${^{1,2,3,4}}8.54$ $\pm$ $0.03$ & ${^{1,2}}21.11$ $\pm$ $0.09$  & ${^{1,2,3,4}}8.80$ $\pm$ $0.09$  & ${^{1,2,3}}21.32$ $\pm$ $0.07$ & $1.3$ \\
\bottomrule
\end{tabular}
}
\caption{Averaged results in WER on the LibriSpeech 100h dataset over 3 runs with standard deviations ($\pm$). SpecAugment with RoBERTa on the target side only (LanguageModel) and with the audio dictionary on the source audio only (AudioDict) already gives consistent relative improvements of 2.7\% to 5.4\% and 5.8\% to 7.0\% respectively across all datasets. 
The language model guided ADA method (ADA-LM) combines RoBERTa and the audio dictionary in an aligned manner and delivers relative improvements of 
11.9\% to 12.0\% on the \texttt{clean} datasets, and of 7.1\% to 8.9\% on the \texttt{other} datasets over the baseline. 
The random token strategy for ADA (ADA-RT) improves even more and gives relative improvements of 18.7\% to 23.5\% on the \texttt{clean} datasets, and of 7.9\% to 9.3\% on the \texttt{other} datasets. 
Prepended numbers denote statistically significant difference to the model numbered in column ``\#'' at the 1\% level. 
}
\label{tab:libsph100}
    \vspace{-6mm}
\end{table*}

\section{Model Architecture}
\subsection{Automatic Speech Recognition} We use the implementation of the Speech-to-Text Transformer by \cite{wang2020fairseq} from the \textsc{fairseq} website 
and add a Connectionist Temporal Classification (CTC) component and the data augmentation code. The model has two convolution layers of stride 2 to down-sample the audio representations by a factor of 4 before the self-attention blocks. There are 12 self-attention layers in the encoder which is followed by 6 layers in the decoder. The embedding dimension is 512, and we set the dimension of feed forward networks to 2,048. In order to achieve faster and more stable convergence, we add another output layer after the encoder so that its parameters are shared between the Cross-Entropy loss $L_{XENT}$ with label smoothing of 0.1 and CTC loss $L_{CTC}$ \cite{watanabe2017hybrid}. The final loss 
is a linear combination of both loss components, $L = \alpha L_{XENT} + (1-\alpha) L_{CTC}$, where we follow \cite{karita2019comparative} and set $\alpha$ to $0.7$. Our implementation is publicly available.\footnote{\url{www.github.com/StatNLP/ada4asr}}

\subsection{ADA Components}
To obtain token-level alignment information between source audio representation and target text, we use the Montreal Forced Aligner \cite{mcauliffe2017montreal}. The alignment information is initially used to construct the audio dictionary as follows. For each training example in the corpus, the construction process iterates over the tokens and adds the aligned audio representations to a key-value store where the key is the token and the value is a pool of aligned audio representations. This audio dictionary 
can be further enriched with pre-calculated speed and frequency perturbations, effectively integrating offline methods in online augmentation.

To get the target token predictions during training under the ADA-LM configuration, we query \texttt{roberta.base}, a pre-trained RoBERTa \cite{liu2019roberta} language model downloaded from the \textsc{fairseq} \cite{ott2019fairseq} examples repository.\footnote{\url{www.github.com/pytorch/fairseq/tree/master/examples/roberta}   (accessed 03/25/2021)} 
We chose the 125M parameter model to trade-off speed and memory consumption.
Under the ADA-RT configuration, tokens are directly sampled from the keys in the audio dictionary in a random manner. 

\section{Experiments}
\label{sec:experiments}
We use LibriSpeech\footnote{\url{www.openslr.org/12} (accessed 03/25/2021)} \cite{panayotov2015librispeech} in our experiments since there exist many baselines in well-defined small to medium and large scale scenarios. For our small to medium scale experiments, we use the split \texttt{train-clean-100} and extract subword units of size 5,000 using SentencePiece \cite{kudo2018sentencepiece}. For the large scale experiments, we combine splits \texttt{train-clean-100}, \texttt{train-clean-360} and \texttt{train-other-500} to form the 960h data and extract 10,000 subword units. We use log Mel-filter banks of 80 dimensions as our acoustic features. We filter data instances which have more than 3,000 frames or are longer than 80 subword units in the training sets. This results in the removal of 50 samples and 78 samples in \texttt{train-clean-100} and \texttt{train-960}, respectively. In both scenarios, the same audio dictionary extracted from \texttt{train-clean-100} is used. 

In training, we use Adam optimizer with a peak
learning rate of 2e-3 and warmup of 12,500 steps. We accumulate the gradients for 8 mini-batches before updating with at most 40,000 frames per mini-batch. The learning rate is adjusted according to the inverse square root learning rate schedule. 
We also use a static mixture schedule for the type of replacement such that a significant amount of transcriptions remains unchanged. 
The static mixture schedule of ADA was tuned on the respective LibriSpeech dev sets.

\begin{table}[h]
\centering
\footnotesize
{
\begin{tabular}{lccccc}
\toprule
{data} & \multicolumn{2}{c}{Aligned Augm.} &\multicolumn{2}{c}{AudioDict} & ADA\\
{set} & {sentences} & {tokens}  & {sentences} & {tokens} & {sentences} \\
\midrule
100h & 50\% &  20\% &  15\% &  20\% & 65\%\\
\midrule
960h & 30\% &  20\% &  21\% &  15\% & 51\%\\
\bottomrule
\end{tabular}
}
\caption{Details of the static mixture schedule of ADA we used in the experiments on LibriSpeech 100h and 960h datasets. }
\label{tab:mixtureschedule}
\vspace{-5mm}
\end{table}

Details of our static mixture schedule are listed in Table \ref{tab:mixtureschedule}. 
In each mini-batch, there is a fraction of sentences augmented using our aligned method and fraction of sentences augmented using the AudioDict only. The ``tokens'' columns indicate the amount of token-level replacements applied to these sentences. %
The final column lists the total of augmented sentences.
In all configurations, we apply SpecAugment at the end with a frequency mask parameter of 30 and a time mask parameter of 40, both with 2 masks along their respective dimension.

\begin{table*}[ht]
\centering
\footnotesize
{
\begin{tabular}{llrrrrc}
\toprule

\# & model & {\texttt{dev-clean}} & {\texttt{dev-other}} & {\texttt{test-clean}} & {\texttt{test-other}} & \texttt{time}\\
\midrule
1 & SpecAugment (baseline) & $3.94$ $\pm$ $0.10$ & $8.47$ $\pm$ $0.13$ & $4.45$ $\pm$ $0.10$  & $8.29$ $\pm$ $0.06$ & $1.0$\\
2 & ADA-LM & $3.78$ $\pm$ $0.02$ & $8.25$ $\pm$ $0.00$ & ${^{1}}4.22$ $\pm$ $0.07$ & ${^{1}}8.07$ $\pm$ $0.01$ & $1.8$\\
3 & ADA-RT & ${^{1,2}}3.50$ $\pm$ $0.03$ & ${^{1}}8.14$ $\pm$ $0.05$ & ${^{1,2}}3.75$ $\pm$ $0.01$ & ${^{1}}7.97$ $\pm$ $0.04$ & $1.3$\\
\bottomrule
\end{tabular}
}

\caption{Averaged results in WER on the LibriSpeech 960h dataset over two runs with standard deviations ($\pm$). Our language model guided aligned ADA method (ADA-LM) is about twice as slow as SpecAugment. ADA-LM gains relative improvements of 4.1\% to 5.2\% on the \texttt{clean} datasets, and of 2.6\% to 2.7\% on the \texttt{other} datasets. ADA-RT gains relative improvements of 11.2\% to 15.7\% on the \texttt{clean} datasets, and of 3.9\% on the \texttt{other} datasets.
Prepended numbers denote statistically significant difference to the model numbered in column ``\#'' at the 5\% level determined following \cite{riezler-maxwell-2005-pitfalls}.
}
\label{tab:libsph960}
    \vspace{-6mm}
\end{table*}

\subsection{Results on train-clean-100} 
On this dataset, our ASR models are trained for 200 epochs and checkpoints are averaged over the last 75 epochs for each setting. We report mean and standard deviation of micro word error rate (WER) calculated over 3 runs on the standard data splits, i.e. \texttt{dev-clean}, \texttt{dev-other}, \texttt{test-clean} and \texttt{test-other}. We also perform an ablation study to investigate the effect of each proposed augmentation. Table \ref{tab:libsph100} summarizes the results.
Our main baseline uses only SpecAugment for data augmentation.
Other SpecAugment baselines that were trained on the same data split report comparable results, e.g. baselines in \cite{luescher2019rwth} and \cite{kahn2020selftraining} are worse than ours while the baselines in \cite{laptev2020you} are very close to ours.

Both non-aligned source side only (AudioDict) and non-aligned target side only (LanguageModel) augmentations show consistent reductions in WER of about 0.5--0.8 points on the \texttt{clean} datasets. 
On the \texttt{other} datasets, however, the source side only AudioDict augmentation gives improvements of about 1.4 points in WER while the target side only LanguageModel augmentation gives lower improvements of 0.6--0.7. 
In both non-aligned experiments, we applied augmentations to 50\% of the sentences and 20\% of their tokens. 

In comparison to the plain SpecAugment baseline, the language model guided ADA-LM reduces WER by 1.25 and 1.38 points on the \texttt{clean} datasets, and by 1.62 and 2.09 points on the \texttt{other} datasets. 
Switching to the random token replacement strategy of ADA-RT gives 
even larger reductions in WER by 1.97 and 2.70 points on the \texttt{clean} datasets (relative improvement 18.7\% to 23.5\%), and by 1.81 and 2.18 points on the \texttt{other} datasets (relative improvement 7.9\% to 9.3\%).
This result is surprising at first as the augmented examples mostly represent ungrammatical sentences. 
On second thought, such examples effectively force the model to rely less on the inherent language model of the ASR system, and put more weight on the plain audio recognition component as conditioning factor, resulting in increased model performance on unseen examples. 
This result is confirmed by a side experiment where randomly replacing tokens on the target side only did not in general improve results over the baseline.

\subsection{Results on train-960} 
On this large dataset, we train our ASR model for 150 epochs and average checkpoints of the last 20 epochs for evaluation. Mean and standard deviation of WER are calculated over 2 runs.
Table \ref{tab:libsph960} summarizes our results. 
In comparison to the SpecAugment baseline, we observe consistent improvements for the ADA-based methods on all datasets. 
For the language model guided ADA-LM method, WER is reduced by 0.22 points on the \texttt{other} datasets corresponding to a relative improvement of 2.6\% and 2.7\%, and WER is reduced by 0.26 points and by 0.23 points on the \texttt{clean} datasets corresponding to a relative improvement of 4.1\% and 5.2\% on \texttt{dev} and \texttt{test}, respectively.

Switching to the random token strategy implemented in ADA-RT gives further improvements. WER is reduced by 0.33 points and by 0.32 points on the \texttt{other} datasets which corresponds to a relative improvement of 3.9\%, and WER is reduced by 0.44 points and by 0.70 points on the \texttt{clean} datasets which corresponds to a relative improvement of 11.2\% and 15.7\% on \texttt{dev} and \texttt{test}, respectively.

For the large 960h dataset, comparable numbers to our baselines can be found in \cite{hu2020sapaugment}. They use a transformer architecture implemented in ESPnet \cite{watanabe-et-al-2018-espnet} and report SpecAugment baseline WER scores that are very close to ours.

\subsection{Training Speed and Model Complementarity}
We report average per-instance training time normalized by the baseline model's training time in the last column of Tables \ref{tab:libsph100} and \ref{tab:libsph960}. 
Adding the audio dictionary for augmentation increases training time by a factor of 1.2 compared to SpecAugment alone. Notably, the largest increase in computational effort is introduced by querying the language model. The target side only LanguageModel experiment increases training time by a factor of 2.0, and our language model guided ADA-LM implementation increases training time by a factor of 2.4 for 100h, and 1.8 for 960h. The random token strategy of ADA-RT is thus significantly faster than its language model guided counterpart, resulting in a training time increase factor of only 1.3. This shows that both ADA variants are well-suited for on-the-fly training, where the latter, ADA-RT, is remarkably efficient. 

\begin{table}[h]
\centering
\footnotesize
{
\begin{tabular}{lll}
\toprule
{model} & \texttt{test-clean} & \texttt{test-other} \\
\midrule
w/o Augmentation & $13.74$ & $31.91$ \\
SpecAugment (SA) only & $11.50$ $(-2.24)$ & $23.50$ $(-8.41)$ \\
ADA-RT w/o SA    & $10.95$ $(-2.79)$ & $30.03$ $(-1.88)$ \\
ADA-RT with SA   & $\ \ 8.80$ $(-4.94)$ & $21.32$ $(-10.59)$ \\
\bottomrule
\end{tabular}
}
\caption{Numbers in ``()'' are the differences in WER to the topmost model which was trained w/o any augmentation method. They illustrate that ADA-RT is complementary to SpecAugment. } 
\label{tab:complement}
\vspace{-5mm}
\end{table}

In a side experiment, we evaluated the complementarity between ADA and SpecAugment and report results in Table \ref{tab:complement}. The numbers in parenthesis nicely illustrate that the contributions of ADA-RT w/o SA and SpecAugment directly add up.

\subsection{Significance Testing} 
Significance testing across different runs is not straightforward. Pairwise tests \cite{GillickCox:89} across different runs of %such
models are problematic as the rejection of the null hypothesis might be based on different initializations and not only on architectural model differences. Bootstrap tests \cite{BisaniNey:04}  might be problematic because of the assumption that the test set is representative of the population distribution, something which is not satisfied if train and test data are from different domains. A non-parametric significance test that only relies on the strategy of stratified shuffling is the permutation test, a.k.a. (approximate) randomization test, dating back to \cite{Fisher:35}. For large samples, this test has been shown to be as powerful as related parametric tests \cite{Hoeffding:52}, and it produces fewer Type-I and Type-II errors than the bootstrap \cite{Noreen:89}.

Thus, to compare our systems we conduct significance tests as follows.
For each transcription reference, we compress the score results of each system's runs to a single average, effectively reducing score variance per example. 
This is valid for WER values case because the reference length is constant.
We then determine $p$-values to decide whether our proposed systems are significantly different to their SpecAugment baseline models and we also compare systems against each other. 
For Librispeech 100h, we conduct in total 10 pairwise comparisons of models, thus we apply a Bonferroni correction to the 1\% level and set $\alpha$ to 0.001 (0.01/10). 
For Librispeech 960h, we conduct 3 pairwise comparisons, thus we set $\alpha$ to 0.0166 (0.05/3) for the 5\% level.
Statistical significance of model differences is then determined using an approximate randomization test: 
we set the number of randomly shuffled runs to 1000, and in each run, we exchange the transcriptions' scores of two models with a probability of 0.5 for each example and calculate the test statistic. 

On Librispeech 100h, both ADA models as well as the LanguageModel and AudioDict models are significantly different to their SpecAugment baseline models on all four datasets. On the \texttt{clean} datasets, the ADA models are also significantly different to all  models that don't use alignment information, with the best performing ADA-RT model being significantly different to ADA-LM.
On Librispeech 960h, significant differences of ADA-LM to their SpecAugment baseline models were identified on \texttt{test-clean} and \texttt{test-other}. 
The ADA-RT method, however, is significantly different to its SpecAugment baseline models on all datasets, and turns out to be also significantly different to ADA-LM on the \texttt{clean} datasets. 

\section{Conclusion}
We proposed a data augmentation method that makes use of alignment information to create effective training examples. An audio dictionary that is extracted from the training set can be queried with low computational overhead and is used to construct previously unseen utterances and speaker combinations. 
By combining textual token replacements with the audio dictionary in an aligned manner, our model is able to construct unseen examples on-the-fly with acceptable impact on training speed if we use predictions from a language model. In case we employ a random strategy for token replacements, we see even larger improvements with very little impact on training speed. 
Our aligned methods show significant improvements in WER over methods that don't use alignment information on small to medium and large LibriSpeech datasets.

In future work, we would like to further strengthen the idea of aligned augmentation. For example, one could replace $n$-grams building semantically meaningful phrases, or apply filters on stopwords or frequency of words.
The audio dictionary could be enriched with audio representations extracted on speed perturbed raw audio waveforms, and entries could be stretched temporally. 
We also plan to investigate methods for combining existing data augmentation techniques such as learning a sample-adaptive policy, or to combine our approach with self-training given source audio data, and we plan to evaluate our methods on noisier datasets and other languages.

\section{Acknowledgements}
This research was supported in part by the German research foundation (DFG) under grant RI-2221/4-1.
We'd like to thank the reviewers for their helpful comments, which we mainly address on our github repository page due to 
length constraints.

\bibliographystyle{IEEEtran}

\bibliography{references.bib}

\end{document}